\documentclass[letterpaper, 10 pt, conference]{ieeeconf}  

\IEEEoverridecommandlockouts                              

\usepackage{svg}
\usepackage{amsmath}
\usepackage{amsfonts}
\usepackage{booktabs,siunitx}

\usepackage{graphicx}
\usepackage{comment}
\usepackage[caption=false]{subfig}

\usepackage{todonotes}

\title{\LARGE \bf
Data-driven Grip Force Variation in Robot-Human Handovers
}

\author{Parag Khanna$^{1}$, Mårten Björkman$^{1}$ and Christian Smith$^{1}$
\thanks{$^{1}$ Division of Robotics, Perception and Learning (RPL), EECS, KTH Royal Institute of Technology, Sweden
        {\tt\small paragk@kth.se,} {\tt\small celle@kth.se,} {\tt\small ccs@kth.se}}%
}

\usepackage{hyphenat}
\makeatletter
\let\NAT@parse\undefined
\makeatother
\usepackage[hidelinks]{hyperref}
\begin{document}

\maketitle
\thispagestyle{empty}
\pagestyle{empty}

\begin{abstract}
Handovers frequently occur in our social environments, making it imperative for a collaborative robotic system to master the skill of handover. In this work, we aim to investigate the relationship between the grip force variation for a human giver and the sensed interaction force-torque in human-human handovers, utilizing a data-driven approach. A Long-Short Term Memory (LSTM) network was trained to use the interaction force-torque in a handover to predict the human grip force variation in advance. Further, we propose to utilize the trained network to cause human-like grip force variation for a robotic giver. 

\end{abstract}

\section{Introduction}
In a handover, the giver holds and carries the object to a suitable, pre-determined handover location while the taker reaches for the object. After this, the giver detects the interaction of the taker with the object and suitably starts decreasing its grip force. The taker forms its grip and increase its grip forces and this is followed by giver completely transferring the object to the taker. A human giver uses multiple sensory inputs, including vision, tactile, and haptic, to start its grip release and accordingly vary its grip force before finally releasing the object.

Many studies (\cite{chan_grip_from_load_second_PR2},\cite{loadshare_strategy-7803296}) show that humans prefer robot behavior that mimics human behavior. Thus, a social robot is expected to lead to human like handovers. Since grip-release is an important part of the handover, a robotic giver should be proficient in it. The robotic giver needs to decide when to start the grip-release, i.e, start decreasing its grip forces and how to decrease them to completely release the object. In this paper, we propose a data-driven approach to compute human-like grip force variation for a robotic giver in handover. We explore the relation between the interaction force-torque and the grip forces in human handovers, utilizing a dataset from a human study \cite{self_humanoids_22_paper}. We further describe the training of a LSTM based architecture which takes in the input as a time-series of the interaction force-torque and outputs a grip force variation for further time-steps. We propose the utilization of this architecture for a robotic giver, which would be able to compute beforehand a human-like grip force variation by sensing the interaction in a robot-human handover scenario.
 

  \begin{figure}[t]
      \centering
     \includegraphics[width=8cm,height=3.3cm,trim={5.2cm 3.5cm 5.2cm 4.70cm},clip]{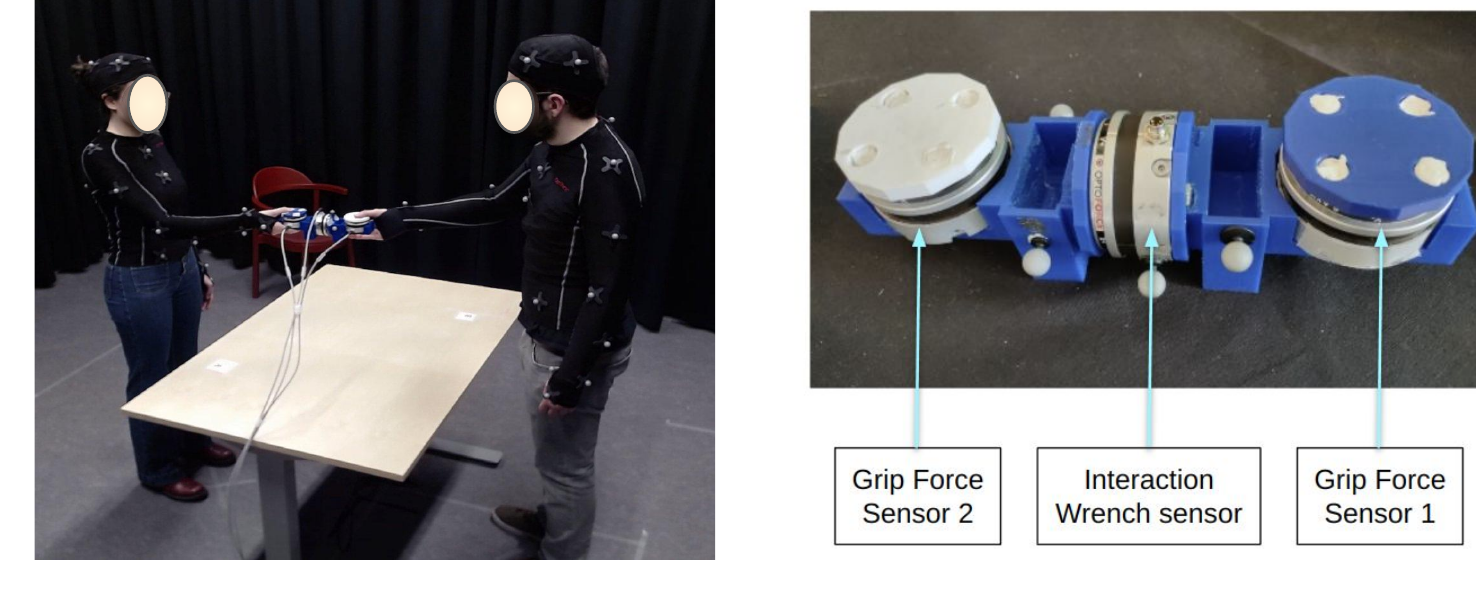}  
       \setlength\abovecaptionskip{-0.4\baselineskip}
      \caption{Study of human handovers and the sensor embedded baton}
      \label{fig:H2H_handover_study}
              \vspace{-4.5mm}

   \end{figure}

\subsection{Related and Prior Work}
500 milliseconds (ms) is the average handover duration according to studies (\cite{chan_grip_from_load_second_PR2},\cite{study_in_person_handover_for_baton}) on grip forces and their variation in human-human handovers. Therefore, if a natural successful handover is desired, a robot giver must make the grip-release decision in a split second.

In \cite{chan_grip_from_load_second_PR2}, a study on human-human handovers involving the vertical passing of a sensor embedded baton was conducted. It was shown that a linear relation exists between load shared and grip force of the human giver. This finding was used in \cite{chan_grip_from_load_second_PR2, chan_grip_from_load_humanoid-6907004} to drive robotic giver's grip force variation during handover depending on load forces in the vertical direction. Their results suggested that human takers preferred the grip-force variation similar to humans. In these handovers, the object was completely released after a slight upward pull force was detected. Based on a human study, the authors in \cite{loadshare_strategy-7803296} modeled grip forces and robot trajectory for handover as a coupled dynamical system based on load sharing. The findings of their human study suggest that humans vary grip force as a function of perceived load share. It is to be noted that the load share is one component of the interaction force.

Humans are indeed the ideal candidates for studying handovers, as we have perfected handovers by doing them in-numerous times. So in a prior work, we conducted a study of human handovers (Fig. \ref{fig:H2H_handover_study}) and proposed a data-driven strategy to command grip-release for a robotic giver \cite{self_humanoids_22_paper}. We studied the interaction wrench ($W$), which includes 3D force-torque, involved in human handovers and related $W$ to the start of human grip-release. Based on human handovers, we trained a LSTM based classifier to classify if an input time series of $W$ corresponded to a grip-release start or not. The trained LSTM was then used with a robotic giver to command its grip release in a robot-human handover experimentation. The robot measured $W$ via a Force/Torque (F/T) sensor placed at its wrist. It was shown that, with the training object, our strategy led to faster grip-release command. It was also ranked most natural
by $95\%$ participants and preferred over other strategies with significance. While the timing of grip-release start is discussed in our prior work, in this paper we discuss how to vary the grip-forces.
 
\section{Human study and Handover Dataset}\label{sec:H2H-study}
\begin{figure}[t]
    \vspace{-3mm}
    \centering
     \subfloat[]{\includegraphics[width=.49\linewidth,height=3.2cm,trim={5.7cm 4cm 6cm 4.9cm},clip]{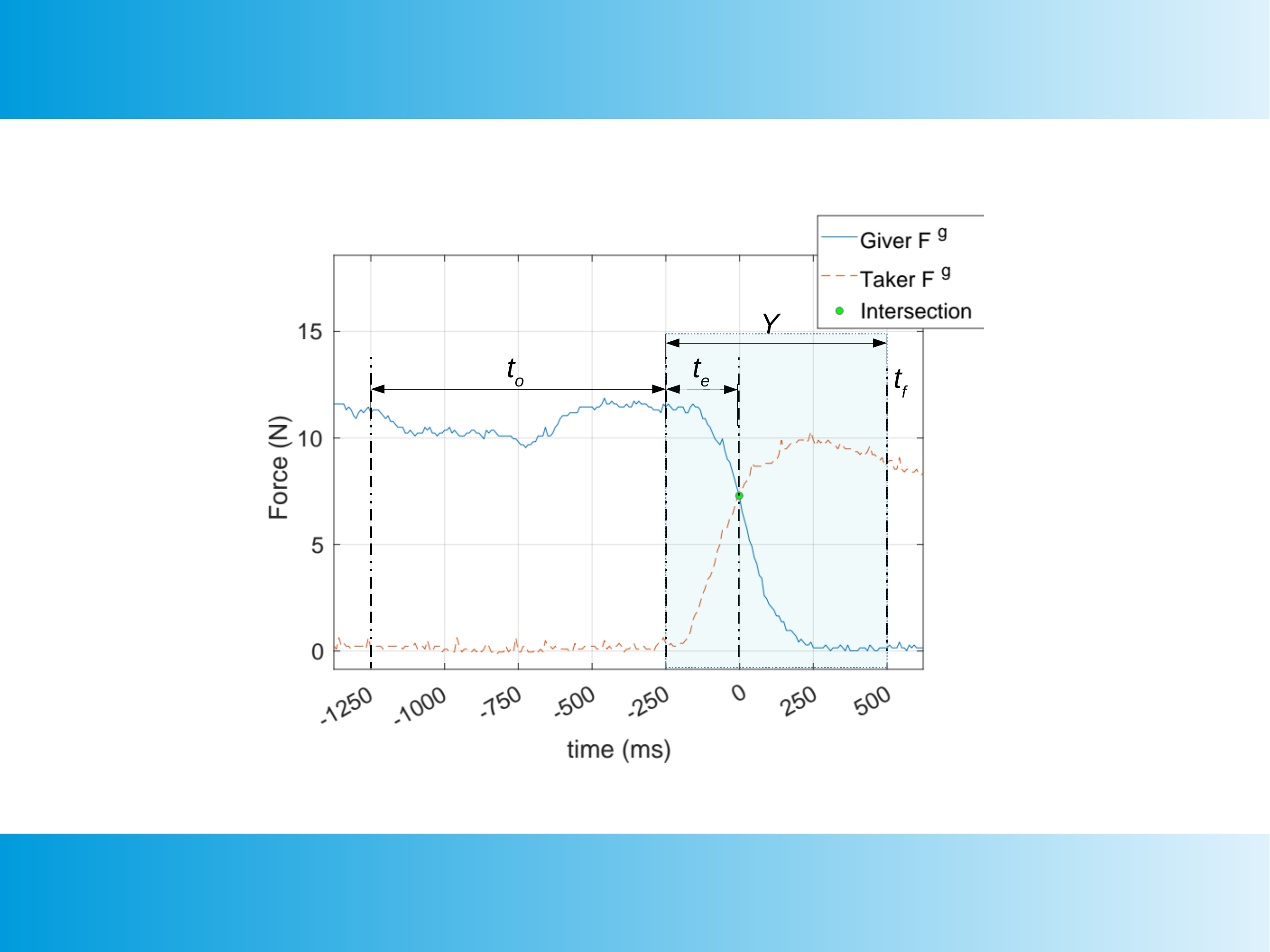}}
    \subfloat[]{\includegraphics[width=.49\linewidth,,height=3.2cm,trim={5.7cm 4cm 5.7cm 4.6cm},clip]{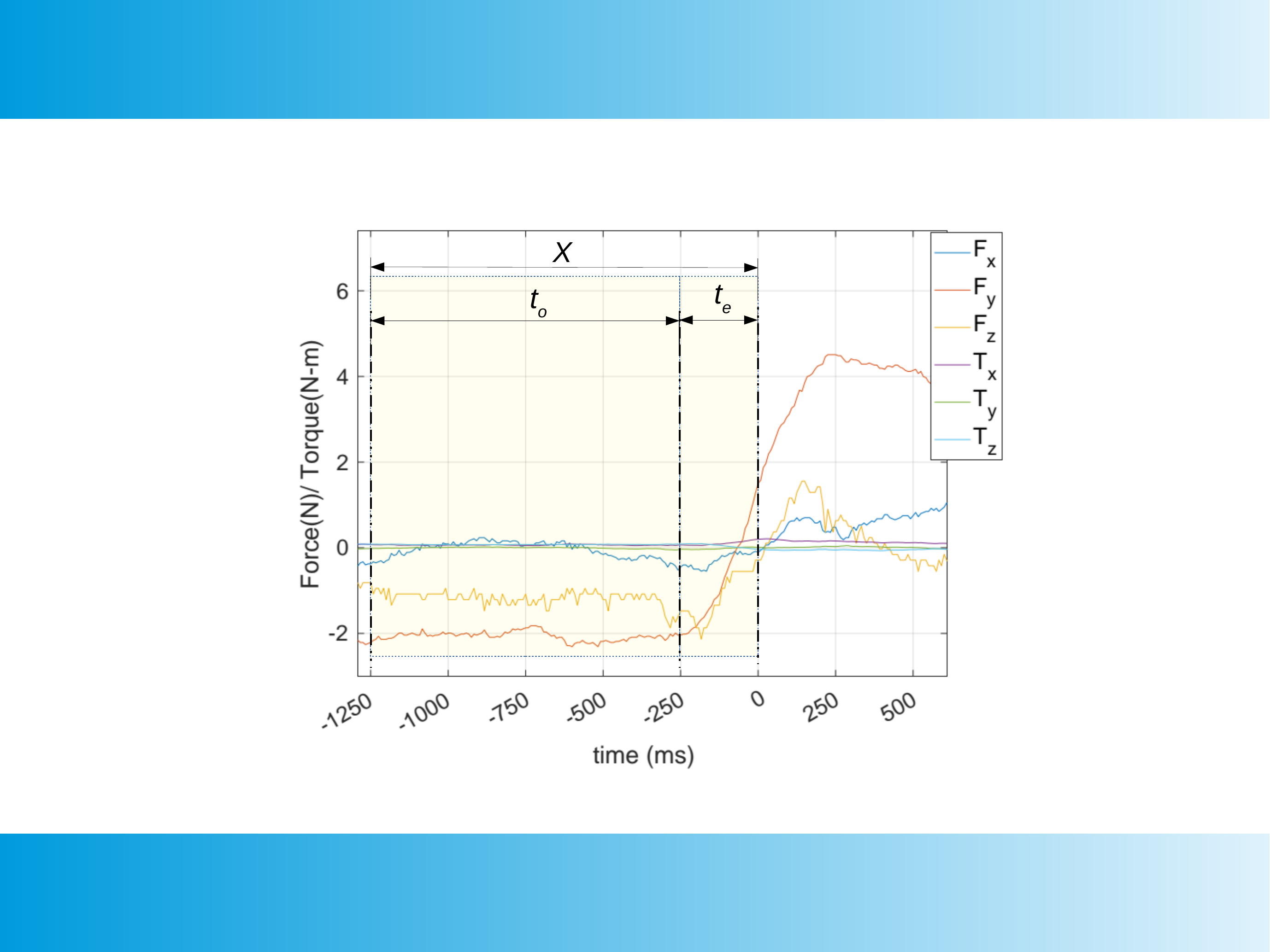}}
    
    
    \caption{Recorded Handover : (a) Grip Forces ($F^g$) (b) Interaction Wrench ($W$): $F$ and $T$ measured for a handover. \textit{X} denote the input time series of \textit{W} for LSTM and \textit{Y} the prediction horizon in giver $F^g$}
    \label{fig:grip_intx_forces_handover}
        \vspace{-5mm}
    \end{figure}

Human handovers were recorded in a study by 13 participant pairs. A sensor embedded baton was transferred to measure the interaction forces and grip forces of the giver and the taker, as shown in the Fig. \ref{fig:H2H_handover_study}. Each handover was separated and is centered about the intersection point of grip forces at $t=0$ ms, as shown in Fig. \ref{fig:grip_intx_forces_handover}(a). 

\section{Learning grip forces variation} \label{sec:Learning from h2h}
 
 All the components of $F$ and $T$ of the interaction wrench $W$ vary across $t=0$ ms shown, being recorded at 120Hz. Also, the grip release action starts before $t=0$ ms and considering the average duration of 500 ms, we choose the region of interest as of $t=[-250 , 250]$ ms. The learning task is to be able to predict the succeeding giver grip forces $F_g$ given a time series of $W$ before $t=0$ ms.
  
\subsection{Dataset and Samples}
 From the saved data for each handover, training samples were taken as time series of the interaction wrench ($W$) starting at $t_o$ and ending at $t_e$ before $t\text{ = }0$ ms, and the corresponding time series of giver grip forces ($F_g$) starting at $t_e$ and ending at $t_f$. The input time series $X$ and the output time series $Y$ are given by the following equation where $W^t$/$F_g^t$ denotes the $W/F_g$ at time $t$.
 \begin{equation}
 \begin{split}
 X(t_o,t_e)= \{W^{t}, t \in [t_o,t_e]\}\\
  Y(t_e,t_f)= \{F_g^{t}, t \in [t_e,t_f]\}.
 \end{split}
 \label{eqn:time_series_Int_wrench}
 \end{equation} 

The LSTM architecture was trained to predict $Y$ given $X$:
 \begin{equation}
 Y = LSTM(X)
\label{eqn:LSTMClassify}
\vspace{-2mm}
\end{equation}
 
\begin{figure}[b]
\vspace{-4mm}
      \centering
        \includegraphics[width=.6\linewidth,height=2.5cm,trim={0.1cm 0.0cm 0.1cm 0.0cm},clip]{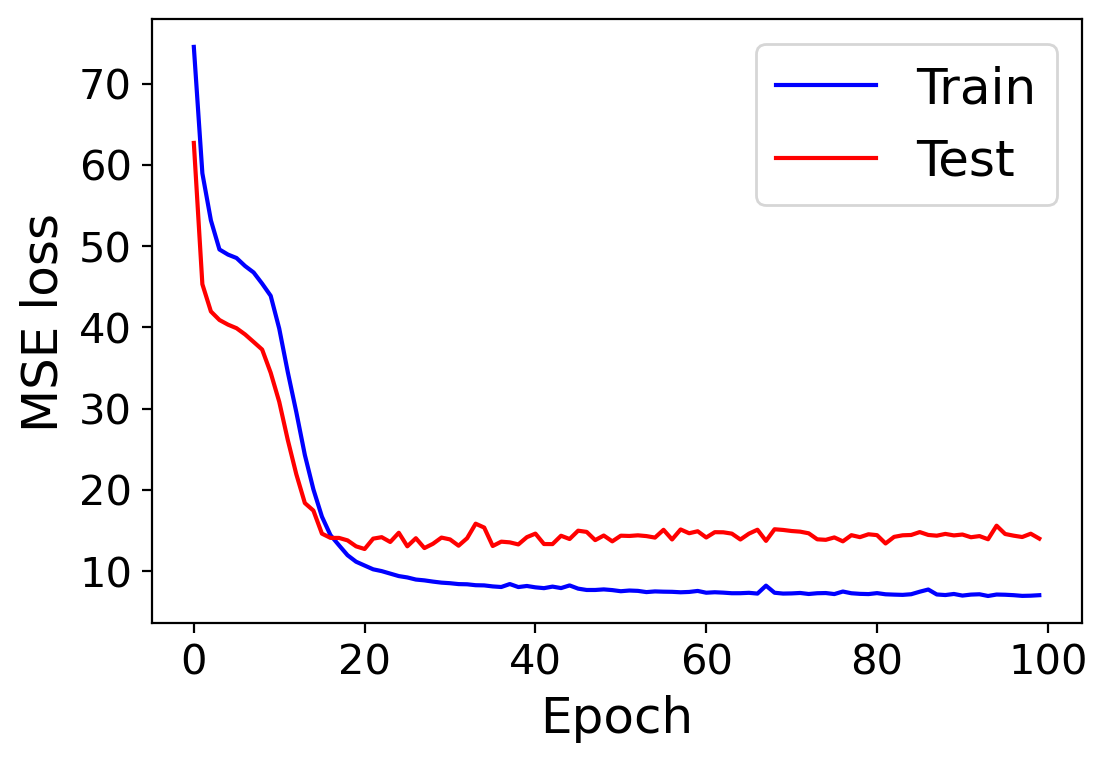}  
     \setlength\abovecaptionskip{-0.2\baselineskip}
    \caption{Loss evaluation for LSTM}
      \label{fig:train_test_loss}
       \vspace{-5mm}
   \end{figure}

To improve robustness, the ending time step for input $X$, $t_e$, was varied between $t=[-250,0]$ ms. The starting time step for the wrench time series $X$, $t_o$,  varied between $t=[-1250,-260]$ ms. This allowed the LSTM to learn predicting the grip forces from a given series of $W$ starting and ending at different instances of sampled handovers, Fig \ref{fig:grip_intx_forces_handover}(b). The prediction horizon for Y was fixed to 583.33 ms (70 time steps at 120Hz), i.e. $t_f=t_e+583.33$ms, for all the samples. This horizon was found sufficient for the giver grip forces to reduce to zero in recorded handovers. 
The training dataset contained samples from recorded handovers of 11 participant pairs leading to a total of 1235 sampled handovers. The test dataset had 253 handover samples from remaining 2 pairs.

 
\subsection{Training and Evaluation} 
\begin{figure}[t]
      \centering
        \includegraphics[width=.95\linewidth,height=3.3cm,trim={0.3cm 10.5cm 0.8cm 3.0cm},clip]{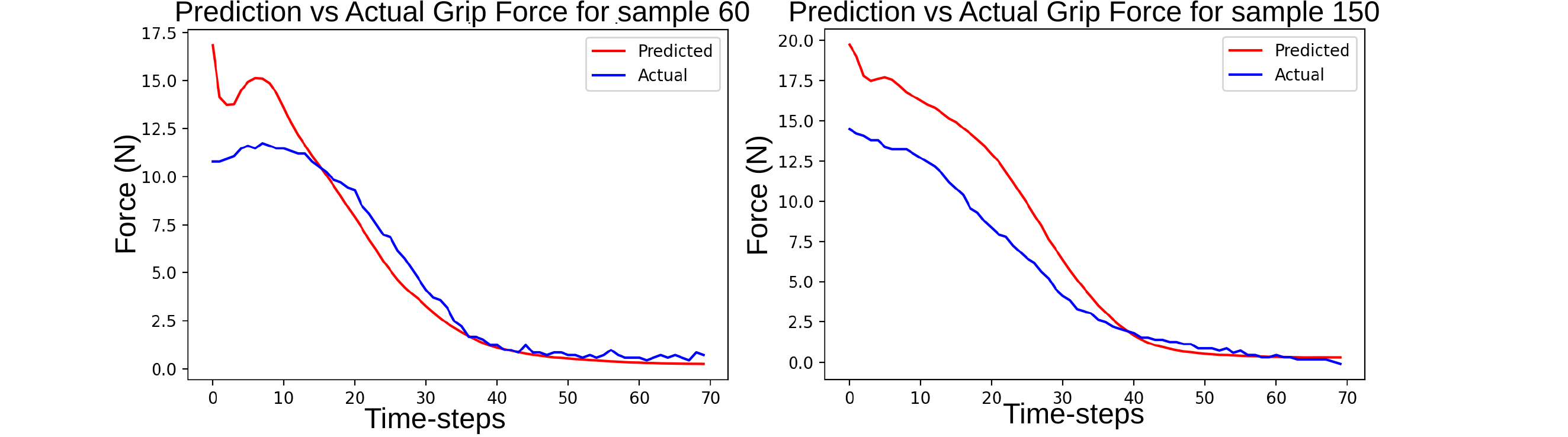}  
     \setlength\abovecaptionskip{-0.3\baselineskip}
    \caption{Comparing predicted grip force to actual grip force of samples from test dataset}
      \label{fig:compare pred actual}
       \vspace{-5mm}
   \end{figure}

The considered architecture contains 2 layers of LSTM with 40 recurrent hidden layers each. The optimized hyperparameters include a learning rate of $5e\text{-}04$ and a batch size of 30. The network was trained on the dataset for 100 epochs, shown in Fig. \ref{fig:train_test_loss}, using mean squared error loss based on difference in predicted and actual grip-forces ($Y$). Adaptive Moment Estimation (Adam) was used for gradient descent optimization. The trained LSTM was used to predict grip force time series given interaction wrench of some samples from test dataset. In Fig. \ref{fig:compare pred actual}, the predicted grip force is compared to actual grip force for typical samples. It is observed that the predicted variation is close to the actual variation, in particular, the forces' value fall to zero together.

Thus, we propose to use the trained LSTM in a robot-human handover, wherein the $W$ is measured by a wrist sensor and once the grip-release command is given, the robot applies a human like grip force variation.

\vspace{0.5mm}

\section{Conclusion and Future work}
We performed a study of human handovers to  investigate the grip forces and the interaction wrench in handovers. We propose and employ a data-driven approach to learn the relation between giver $F^g$ and $W$ and train a LSTM to predict succeeding $F^g$ variation using $W$. For robot-human handovers, we propose that the robotic giver can use the trained LSTM to plan human-like grip force variation in advance according to sensed interaction. For future work, we plan to apply our method to a robotic system tasked with giving objects to a human. It will be interesting to compare our method against grip force variation based on only loadshare which is one component of $W$.

\addtolength{\textheight}{-1cm}   




\section*{ACKNOWLEDGMENT}
This work was in part financially supported by Digital Futures at KTH. 
\vspace{3mm}
\vspace{-1.1mm}
\bibliographystyle{IEEEtran}
\bibliography{IEEEabrv,pk_ref}
\end{document}